\documentclass[conference]{IEEEtran}
\IEEEoverridecommandlockouts
\usepackage{placeins} 
\usepackage{float}
\usepackage{cite}
\usepackage{amsmath,amssymb,amsfonts}
\usepackage{algorithmic}
\usepackage{graphicx}
\usepackage{textcomp}
\usepackage{xcolor}
\usepackage{utfsym}
\usepackage{multirow}
\usepackage{csquotes}

\def\BibTeX{{\rm B\kern-.05em{\sc i\kern-.025em b}\kern-.08em
    T\kern-.1667em\lower.7ex\hbox{E}\kern-.125emX}}
\begin{document}

\title{FLAF: Focal Line and Feature-constrained Active View Planning for Visual Teach and Repeat\\
\thanks{$\dagger$Corresponding author (hzhang@sustech.edu.cn)

Changfei Fu and Hong Zhang are with the Shenzhen Key
Laboratory of Robotics and Computer Vision, Southern University of
Science and Technology (SUSTech), and the Department of Electrical and
Electronic Engineering, SUSTech, Shenzhen, China. Changfei Fu and Wenjun Xu are also with the Peng Cheng National Laboratory, Shenzhen, China. Weinan Chen is with the Biomimetic and Intelligent Robotics Lab, Guangdong University of Technology, Guangzhou, China. This work was supported by the Shenzhen Key Laboratory of Robotics and Computer Vision (ZDSYS20220330160557001).}
}

\author{
\IEEEauthorblockN{ Changfei Fu, Weinan Chen, Wenjun Xu, and Hong Zhang$\dagger$}
}

\maketitle

\begin{abstract}
This paper presents FLAF, a focal line and feature-constrained active view planning method for autonomous orientation adjustment of a rotatable active camera during mobile robot navigation. FLAF is built on a visual teach-and-repeat (VT\&R) system, which enables robots to cruise various paths that fulfill many daily autonomous navigation requirements. The VT\&R system integrates Visual Simultaneous Localization and Mapping (VSLAM) with trajectory following. However, tracking failures in feature-based VSLAM, particularly in textureless regions common in human-made environments, poses a significant challenge to real-world VT\&R deployment. To address this, the proposed view planner is integrated into a feature-based VSLAM system, creating an active VT\&R solution that mitigates tracking failures. Our system features a Pan-Tilt Unit (PTU)-based active mounted on a mobile robot. Using FLAF, the active camera-based VSLAM (AC-SLAM) operates during the teaching phase to construct a complete path map and in the repeating phase to maintain stable localization. FLAF actively directs the camera toward more map points to avoid mapping failures during path learning and toward more feature-identifiable map points while following the learned trajectory. Experimental results in real scenarios show that FLAF significantly outperforms existing methods by accounting for feature identifiability, particularly the view angle of the features. While effectively dealing with low-texture regions in active view planning, considering feature identifiability enables our active VT\&R system to perform well in challenging environments.
\end{abstract}

\vspace{-1mm}
\section{Introduction}
Learning to cruise a path while traversing it is a fundamental capability for mobile robots\cite{b1}. Considering that humans and vehicles mainly rely on various flexibly fixed paths to repeatedly shuttle between multiple locations, teach and repeat (T\&R)\cite{b2} is an essential technique for robots to learn to navigate the paths that cover a major part of autonomous navigation requirements. This technique can support many robotic applications, such as household robots traveling between different rooms\cite{b3}, delivery robots taking goods from the logistics center to the target building\cite{b4}, and autonomous buses following a mostly fixed trajectory.

As a type of natural visual sensor, monocular cameras are cost-effective, energy-efficient, and versatile, making them suitable for a wide range of environments and applications. The T\&R approaches that predominantly utilize visual sensors are referred to as visual teach-and-repeat (VT\&R)\cite{b1}, which is a significant motivation of the research in visual simultaneous localization and mapping (VSLAM)\cite{b5}. During teaching, the robot reconstructs the surrounding landmarks by VSLAM while traversing a path under guidance\cite{b6}. In the repeating phase, the previously saved path map is reloaded to localize the robot for navigating the taught trajectory.


\begin{figure}
  \centering
  \includegraphics[width=1\linewidth]{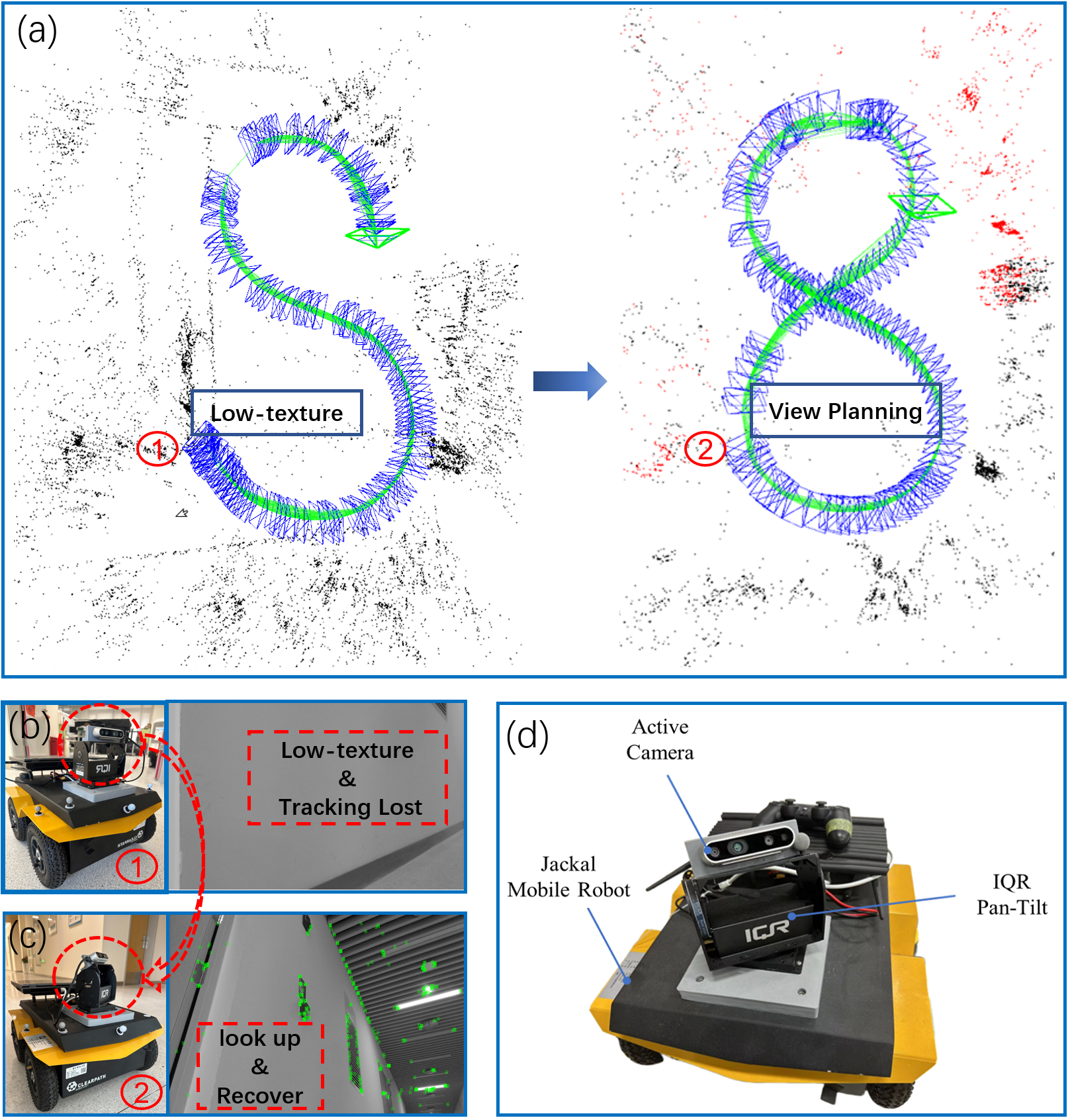}
  \vspace{-6mm}
  \caption{ (a) shows the failure of the passive VT\&R system to learn a complete path due to a low-texture region, which is overcome by our active VT\&R with view planning. The orientations and views of the fixed camera are depicted in (b). In contrast, with our FLAF-constrained view planning and an active camera, the active VT\&R system successfully navigates this challenging path by directing the camera toward feature-rich regions, as shown in (c). (d) presents our mobile robot equipped with a PTU-based active camera. }
  \label{fig1}
\vspace{-7mm}
\end{figure}

Although feature-based VSLAM systems\cite{b7,b8} achieve impressive robustness and stability in large indoor and outdoor environments\cite{b9}, they all suffer from the view angle-dependent affine changes\cite{b10} of features, and the map for real-time localization are usually too sparse for complex applications\cite{b11} that need dense reconstruction. Fortunately, the limited view-angle invariance of features and the sparse map for reliable localization are enough for path following in VT\&R. Using a passive VSLAM system, we achieved a high success rate and reliability in passive VT\&R (using a fixed camera) with feature-based monocular VSLAM. However, tracking failures\cite{b12} caused by textureless regions in human-made environments are still limiting VSLAM and VT\&R to be used in the real world. To solve this, existing active SLAM methods\cite{b12,b26} mostly choose to change the robot trajectory, which is not suitable for VT\&R. To actively select informative views without interfering in the trajectory, our active VT\&R system integrates a pan-tilt unit (PTU)-based active camera with feature-based VSLAM (Fig. \ref{fig1}).

Our motivation is to design a feature-based VT\&R system coupled with active view planning for tracking failure avoidance. Previous active camera-based VSLAM 
(AC-SLAM) methods\cite{b15,b16,b17} have primarily focused on maintaining stable localization during the mapping process but have not demonstrated how to effectively reuse the map with active view planning for navigation tasks. Compared to the mapping process when the active camera has no choice but to orient to the newly built map points, there are more choices of map points when the map is complete during navigation. While the robot is moving forward in the direction indicated by the arrows (Fig. \ref{failure case}), existing view planning methods \cite{b15,b16,b17} that focus the active camera on the regions dense with map points may fail in some VT\&R cases because they do not consider the feature identifiability (the ability of the feature detector to identify a 3D map point). These map points targeted by the active camera may be triangulated by earlier keyframes taken from viewpoints significantly different from the current one, making them unidentifiable by the feature algorithms due to their substantially different visual appearances\cite{b26}. To address this, we have designed an active camera-based VT\&R system featuring an innovative active view planning method that accounts for the view angles of map points. Our FLAF-based view planner is focal line-centric, as its direction dictates the angles of the active camera.

In this paper, we present a feature-based active VT\&R system incorporating an active camera and a novel active view planning method. The main contributions are as follows: (1)We propose a focal line and feature (FLAF)-constrained view planner that addresses failures of visual repeating by accounting for the angle difference between the current viewpoint and those
at which the map points were triangulated. (2) We integrate active view planning into passive VT\&R to build up the active VT\&R by resolving the robot poses from camera localization and PTU angles as the input of passive VT\&R. To the best of our knowledge, this is the first demonstration of the VT\&R system coupled with active view planning. (3) Our code is publicly available at: https://github.com/Changfei-Fu.

\section{RELATED WORK}
Our active VT\&R builds on the VSLAM system tightly coupled with active view planning for tracking failure avoidance.

\subsection{Visual Teach and Repeat}

\begin{figure}
  \centering
  \includegraphics[scale=0.15]{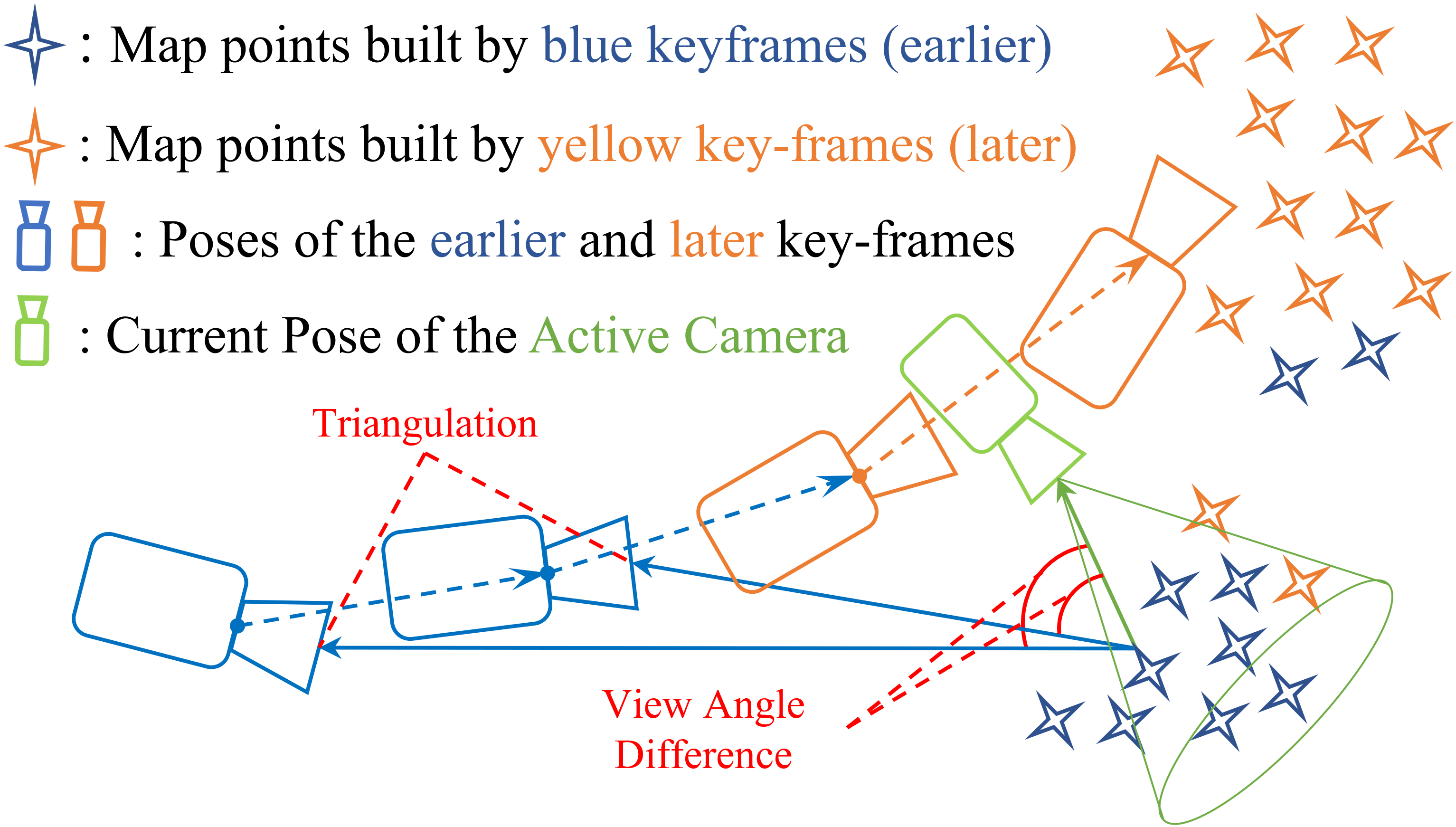}
  \vspace{-2mm}
  \caption{ The failure case of existing view planning methods that consider only the number of map points within the Field of View and their distance to the camera center. The keyframes are selected by the VSLAM system during teaching. During repeating, although the blue points are denser and closer to the active camera, they were triangulated from significantly different viewpoints of earlier keyframes, making them difficult to identify from the current view due to appearance change. }
  
  \label{failure case}
  \vspace{-6mm}
\end{figure}

The classical work of VT\&R\cite{b1} using a feature-based stereo VSLAM was later developed as VT\&R2\cite{b2}, which utilizes multiple taught experiences to address environmental appearance changes. Despite the significant progress, this series of works all suffer from the tracking failure caused by low-texture regions\cite{b12} as they all rely on fixed cameras and feature-based VSLAM. Based on the well-established VT\&R2\cite{b2}, Warren et al.\cite{b18} build a gimbal-stabilized VT\&R system in which the gimbal is passively utilized to stabilize the camera or manually steered to avoid degeneracy in the teaching phase. In the repeating phase, the camera actively rotates to the nearest keyframe in the taught graph. Although this work justified the necessity of an active camera in VT\&R, the gimbal is manually steered in the teaching phase instead of autonomous operation. Previous works designed various active view planning methods for AC-SLAM\cite{b15,b16,b17,b18}. However, they fail to account for the appearance change of the same feature observed from different view angles. Similar to the recently proposed uncertainty-driven view planning (UDVP)\cite{b17}, these approaches tend to rotate the camera toward nearby map points, leading to frequent failures during repeating for they ignore the affine change of features\cite{b10} (Fig. \ref{failure case}). To solve these problems, we implement the same autonomous view planning method (FLAF) in both phases of VT\&R.

Confronted with the aforementioned tracking failure and occlusion, Mattamala et al. \cite{b14} designed a VT\&R system that allows the quadruped to alternate between multiple cameras mounted at different positions on the robot. However, each one of the cameras still encounters the issue of tracking failure. Each camera builds several sub-maps, and ensuring their completion and consistency during merging is difficult without effectively addressing the challenging views. Our proposed AC-SLAM VSLAM can enhance the performance of each camera in this system\cite{b14}. Additionally, our VT\&R system is designed more intuitive and compact, utilizing fewer computational resources.

\subsection{Visual Simultaneous Localization and Mapping}

Our VT\&R system consists of a 3D reconstruction module for the mobile robot to remember a traversed path. In the teaching phase, the robot learns the landmarks in feature-based VSLAM. With the input of an image set that shares observations of the environment, the task of estimating camera motions and a geometrical reconstruction is called Structure From Motion (SfM)\cite{b19}. For a mobile robot with a moving video camera, a system that performs SfM for every image as it is captured is called real-time SfM or VSLAM. Specifically, our proposed active view planning method is designed according to the local map built by VSLAM systems\cite{b9}.

The most popular implementations of VSLAM\cite{b7,b9,b8} follow the principle of aligning the current image to the already-built map for camera localization. Bundle adjustment (BA) based on data association between the current frame and keyframes is conducted locally and globally to obtain a consistent map\cite{b20}. To use time-consuming bundle adjustment in SLAM for optimizing a consistent map, a hierarchical optimization strategy is proposed with the concept of window and local bundle adjustment\cite{b21}. 

In \cite{b9}, a sophisticated local map is designed to align the current frame and implement the local BA. Motion estimation by aligning the current image to the local map is called local map tracking\cite{b7}. Deng\cite{b12} et al. identify a particular tracking failure caused by the incapability of associating enough features. In \cite{b12}, the authors also indicate that the likelihood of tracking failure approximates zero if the number of associated map points exceeds a certain threshold.

\subsection{Active View Planning for VSLAM}

The seminal work in active view planning for VSLAM is presented in \cite{b16}, where Davison et al. demonstrate that the key to active vision for VSLAM is the continual views on a succession of features, along with determining the moment to explore new features. Following this concept, \cite{b15} suggests observing the unexplored areas and repositioning the active camera when there is an insufficient number of features for localization. However, this approach does not ensure precise localization, as a view direction identical to the initial one may not reproduce the original image due to the robot's movement. 

\begin{figure}
  \centering
  \includegraphics[scale=0.14]{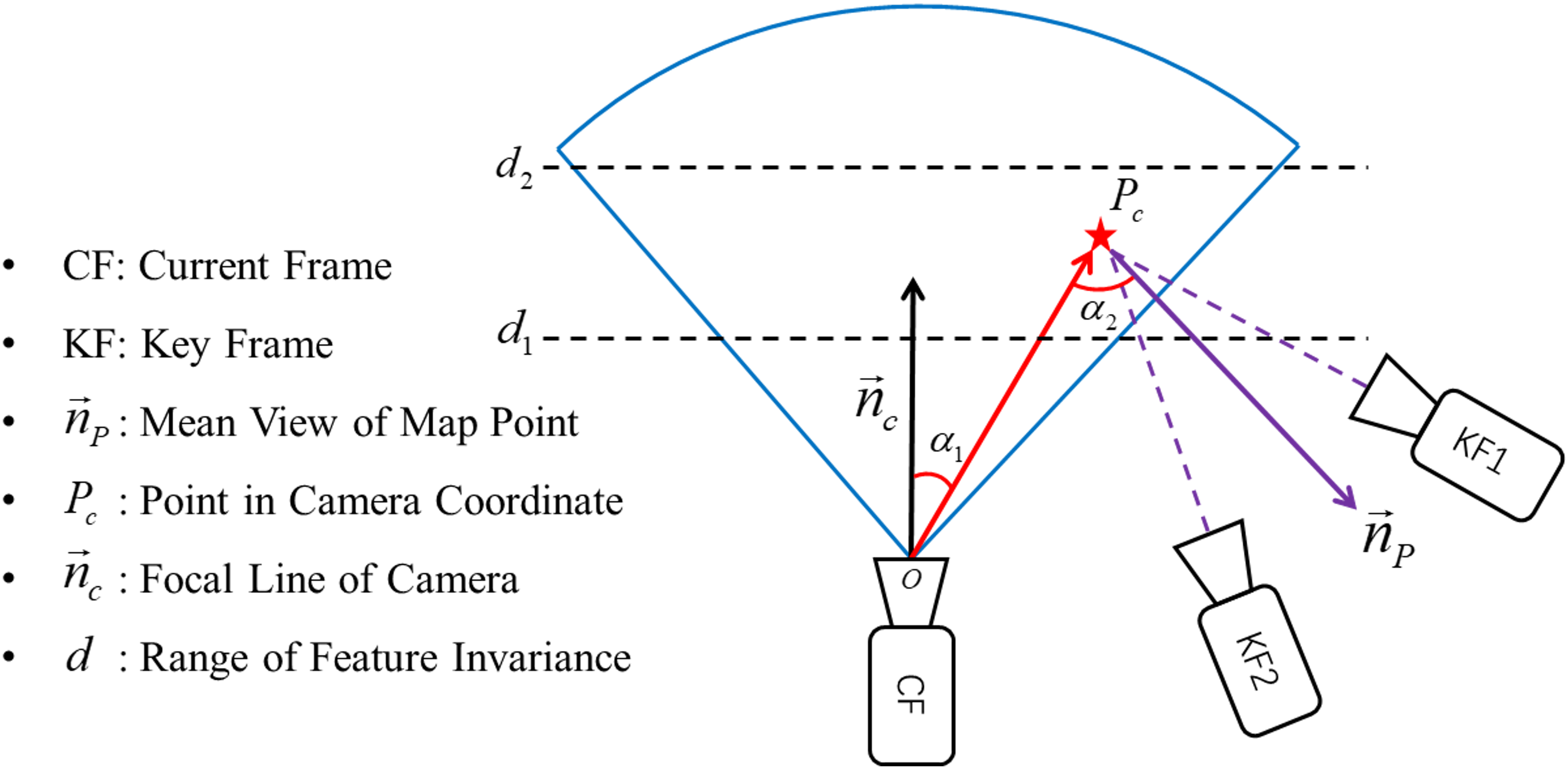}
  \vspace{-2mm}
  \caption{The observation model of FLAF is used to evaluate the camera pose relative to a map point (3D coordinate). For each pair of PTU angles, the scores of all map points in the local map are summed to assess the view direction. In this model, $d_{1}$ and $d_{2}$ represent the invariant distance range of a feature point, which constrains the distance between the map point and the optical center. The angle $\alpha_{1}$ denotes the angle between the line of sight to the map point and the current camera focal line, while $\alpha_{2}$ represents the angle between the line of sight to the point and its mean view line as captured by multiple keyframes. These three metrics are combined to evaluate the camera poses with respect to the local map, determining the optimal PTU angles. }
  \label{FLAF}
  \vspace{-5mm}
\end{figure}

In active camera-based VSLAM, the primary challenge before exploring new features is maintaining stable and accurate localization\cite{b16}. In \cite{b17}, Warren et al. propose the UDVP view planner to assess the pan-tilt angles relative to map points. The UDVP view planner performs well in capturing more map points within the camera's Field of View (FoV). However, as shown in Fig. \ref{failure case}, it leads to tracking failure by orienting at unrecognizable map points during VT\&R. Our FLAF view planner improves upon previous methods by the observation model shown in Fig. \ref{FLAF}, which accounts for not only the view angle between the camera's focal line and the light path of a map point but also the angle between the light path and the normal of map point, ensuring both the number of map points and feature identification. 

The strategy to consider both $\alpha_{1}$ and $\alpha_{2}$ shown in Fig. \ref{FLAF} was initially introduced for feature selection in the VSLAM method described in \cite{b9}, which forms the foundation of our active VT\&R. This strategy is also employed in the active SLAM method presented in \cite{b12}, where a fixed camera is used for navigation and exploration. Similar to \cite{b9}, the method in \cite{b12} defines specific ranges of distance and $\alpha_{2}$ (see Fig. \ref{FLAF}) to screen map points within the camera's FoV. The  \enquote{FLAF without scoring} method in our comparative experiments can be seen as an implementation of active VT\&R using the view planner in \cite{b12}, despite the differences in the robot and task. It is worth noting that Mostegel et al.\cite{b26} identified several metrics for feature recognition and validated the effectiveness of using a cosine function to estimate the likelihood of feature identification from different view angles.

\vspace{-2mm}

\section{Approach}

\subsection{System Overview}

\vspace{-5mm}

\begin{figure}[H]
  \centering
  \includegraphics[scale=0.25]{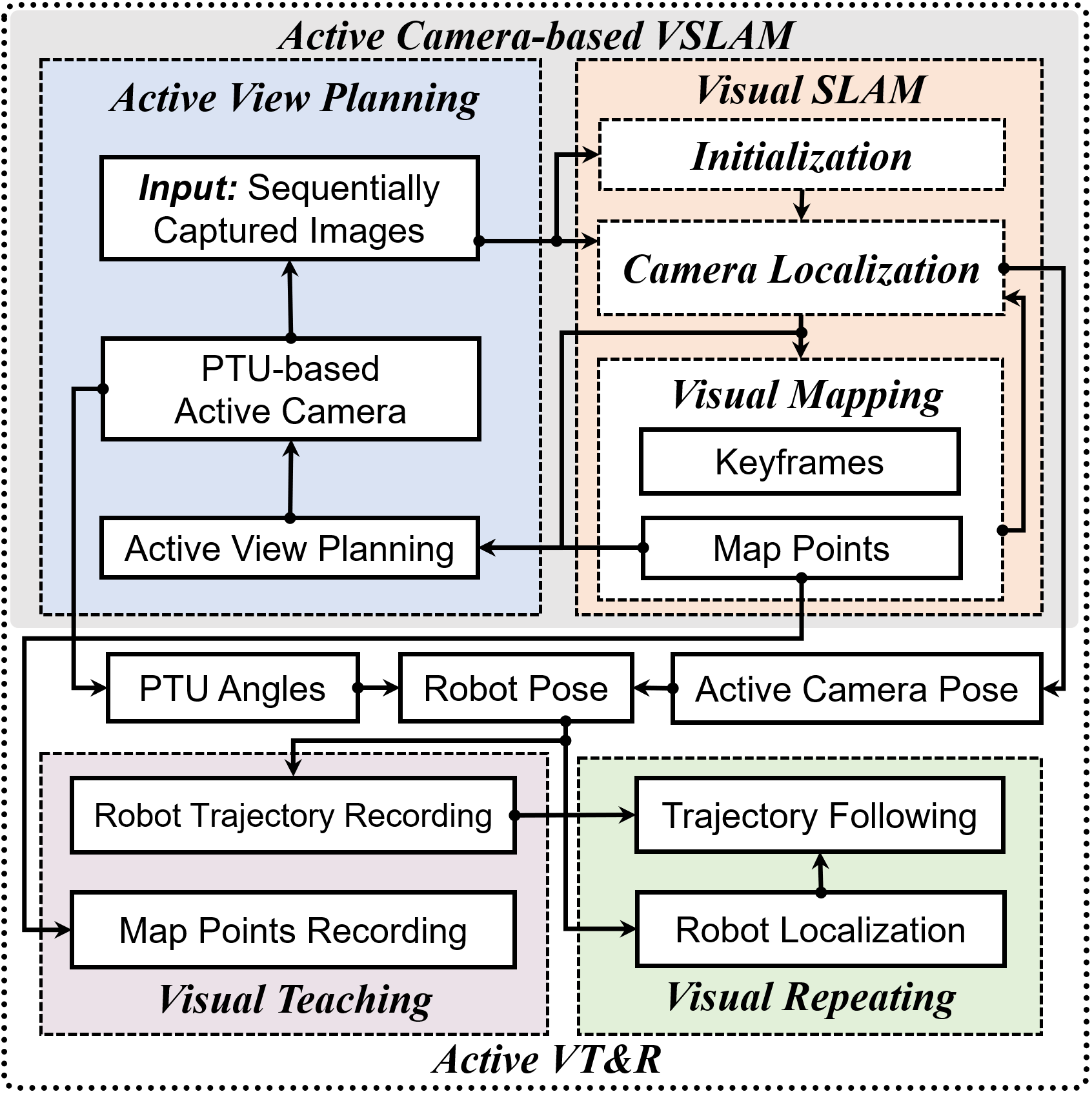}
  \vspace{-3mm}
  \caption{Active VT\&R pipeline, showing the relationship between VSLAM, view planning, and active VT\&R. In both phases of VT\&R, the AC-SLAM is performed and the PTU rotates automatically based on real-time perception, while the mapping module is deactivated during repeating.}
  \label{framework of active VT&R}
  \vspace{-3mm}
\end{figure}

As shown in Fig. \ref{framework of active VT&R}, our active VT\&R system is built on the AC-SLAM, which integrates ORB-SLAM2\cite{b9} and our FLAF-constrained active view planning. Our method for active camera-based path following resolves the robot poses from camera localization and pan-tilt angles, as the input of VT\&R. During both phases of VT\&R, the AC-SLAM operates continuously, with the camera orientation automatically adjusted in real-time based on perception feedback. The path map and robot trajectory created during teaching are reloaded before repeating, with the mapping module deactivated. The robot is initially placed near the taught path with a similar orientation to the taught one. Once the repeating begins, the robot autonomously cruises along the taught path. During repeating, the AC-SLAM provides the path-following module with trajectory reference and localization service, using the previously stored map.

\subsection{VSLAM with Active Camera}
\begin{figure*}
  \centering
  \includegraphics[scale=0.27]{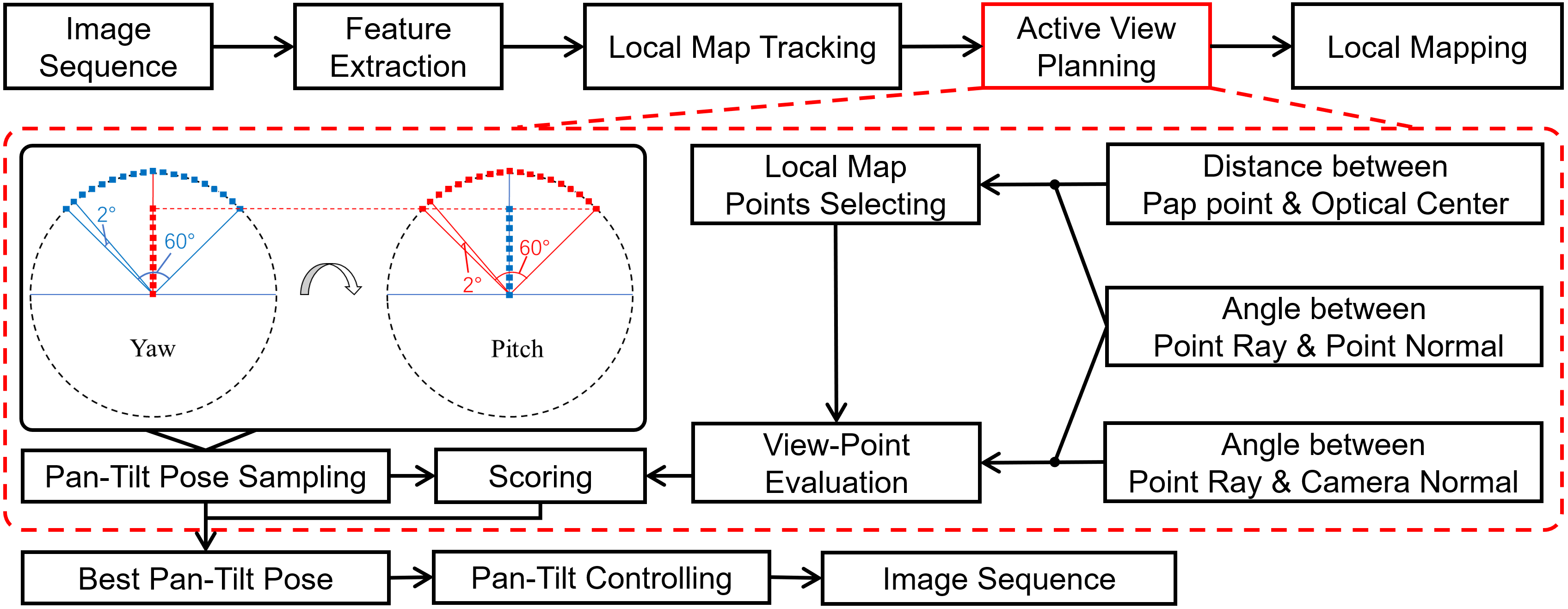}
  \vspace{-2mm}
  \caption{The implementation of sampling-based view planning, which is inserted between the local map tracking and the local mapping of the passive VSLAM system. The PTU angles in the range of 60$^\circ$ for both yaw angle and pitch angle are sampled with a set interval of 2$^\circ$ and scored by our FLAF-constrained view planner. The pan-tilt sample with the highest score, as determined by FLAF, is sent to the PTU control module to adjust the camera direction accordingly.}
  \label{framework of active SLAM}
  \vspace{-5mm}
\end{figure*}

In passive VSLAM, the next image input is determined by the camera movement. To avoid low-texture views, we insert (Fig. \ref{framework of active SLAM}) the feature-based active view planning between local map tracking and local mapping. During both phases of T\&R, the AC-SLAM is performed, and the active camera rotates automatically based on real-time perception. The camera orientation, in turn, dictates the input images for the VSLAM.

While the camera is moving, images are sequentially captured and input to the AC-SLAM. For the k-th input image $\mathbf{I}_{k}:\mathbb{R}^{2}\rightarrow{\mathbb{R}}$, ORB features\cite{b22} are extracted for local-map tracking, where features are aligned with the local map to estimate current camera pose $\mathbf{X}_{k,w}\in \rm{SE(3)}$. If sufficient new features are found, this frame is decided as a keyframe and these features are triangulated into the map by local keyframes. This process of projecting new keyframes and features into the map coordinate is called local mapping. A local map including a network of keyframes $\big\{\boldsymbol{F}_{i}\big\}$ connected by feature matching and their associated map points $\big\{\mathbf{P}_{j}^{i}|\mathbf{P}_{j}^{i}\in\mathbb{R}^{3},j=0,1,2,...,m_{i}\big\}$ are denoted with:
\begin{equation}
       \boldsymbol{M}_{l}=\big\{\mathbf{P}_{0}^{i},\mathbf{P}_{1}^{i},...,\mathbf{P}_{m_{i}}^{i},\boldsymbol{F}_{i}|i=0,1,2,...,n_{i}\big\}
\end{equation}
where $m_{i}$ is the quantity of the map points associated with the i-th keyframe and $n_{i}$ is the quantity of the keyframes in the local map. A map point only refers to a 3D coordinate, and its descriptor can be computed from the keyframe by which this point is triangulated. According to the descriptor matching, keyframes that share observations with $\mathbf{I_{k}}$ and their associated map points make up the local map. 

For any $\mathbf{P}_{j}^{i}$ in $\boldsymbol{M}_{l}$ that is within the current FoV, if it can be matched with an ORB feature at coordinate $\mathbf{p}_{j}^{i}\in\mathbb{R}^{2}$ on $\mathbf{I}_{k}$, it is reprojected onto the $\mathbf{I}_{k}$ by pinhole camera model $\pi:\mathbb{R}^{3}\rightarrow\mathbb{R}^{2}$. The reprojection error as in Eq. \eqref{reprojection error} is applied for optimizations in VSLAM:
\begin{equation}
    \mathbf{e}_{i,j}=\mathbf{p}_{j}^{i}-\pi(\mathbf{P}_{j}^{i})
    \label{reprojection error}
\end{equation}
We use a set $\boldsymbol{S}=\big\{\mathbf{P}_{j}^{i}|\mathbf{P}_{j}^{i}$ can be identified by $\mathbf{I}_{k}\big\}$ to represent the map points in $\boldsymbol{M}_{l}$ that can be identified by the current view. Then local map tracking is conducted by minimizing the cost function\cite{b9}:
\begin{equation}
    \mathbf{X}_{k}^{*}=\arg\underset{\mathbf{X}_{k}}\min\underset{i,j,\\\mathbf{P}_{j}^{i}\in S}{\sum} 
    \mathbf{e}_{i,j}^\mathsf{T}\Omega_{i,j}^{-1}\mathbf{e}_{i,j}
    \label{e4}
\end{equation}
where $\mathbf{X}_{k}^{*}\in \rm{SE}(3)$ represents the pose estimation of $\mathbf{I}_{k}$. 

After local map tracking, a sampling-based optimization (Fig. \ref{active optimization}) of the next best view is adopted based on the local map and real-time localization. The pan-tilt angles are sampled as $\mathbf{q_{s}}=(pan, tilt)$ and transformed into $\mathbf{T}_{pt}(\mathbf{q_{s}})\in \rm{SE}(3)$ to obtain the corresponding sample of camera pose $\mathbf{X}_{S}$, which is directly scored by view planners at a set frequency:
\begin{equation}
    \mathbf{X}_{S} = \mathbf{T}_{pt}(\mathbf{q_{s}}) \ast \mathbf{X}_{k}
\end{equation}
The best sample of pan-tilt angles is sent to the PTU to rotate the active camera. Thus we achieve rotating the active camera on real-time perception feedback.

\subsection{Active View Planning for Feature-based VT\&R}
To implement FLAF by sampling-based optimization, we adopt three metrics indicated in Fig. \ref{FLAF} for sample evaluation. For each pan-tilt sample and map point, one distance and two angles are measured to determine the next best view:
\begin{itemize}
    \item Reward the pan-tilt sample that places the  map point within the feature-invariant distance range of $(d_{1}, d_{2})$ relative to the camera.
    \item Reward smaller $\alpha_{1}(\mathbf{X}_{S}, \mathbf{P}_{j}^{i})$ shown in Fig. \ref{FLAF} which refers to the angle between camera's focal line $n_{c}$ and light path $OP$ of the point. 
    \item Reward smaller $\alpha_{2}(\mathbf{X}_{k}, \mathbf{P}_{j}^{i})$ shown in Fig. \ref{FLAF} which refers to the angle between mean view line $n_{p}$ and light path $OP$ of the point.
\end{itemize}
According to these three principles, we evaluate every pan-tilt sample with $\boldsymbol{M}_{l}$. At first, we eliminate the points that have a distance out of the range $(d_{1}, d_{2})$ or have an angle $\alpha_{2} > 60^\circ$. The rest of the points in the local map make up a collection denoted $\boldsymbol{S}_{r}$. For every map point $\mathbf{P}_{j}^{i}\in \boldsymbol{S}_{r}$, we calculate the score of a pan-tilt sample $\mathbf{q}$ and optimize it by:
\begin{equation}
    \mathbf{q}^*=\arg\underset{\mathbf{q}}{\max}\underset{i,j,\\\mathbf{P}_{j}^{i}\in \boldsymbol{S}_{r}}{\sum}{cos(\alpha_{1})cos(\alpha_{2})}
    \label{active optimization}
\end{equation}
The maximum of each cosine function is achieved when the angle is 0, indicating that the best view angle is obtained when the focal line of the camera overlaps with the mean view direction of the map point.

To explain our design, we denote the number of map points within the FoV that can be matched with an ORB feature in $\mathbf{I}_{k}$ as follows:
\begin{equation}
    N_{S}=f\left(\mathbf{X}_{k},\boldsymbol{M}_{l}\right)
    \label{e5}
\end{equation}
which only indicates that $N_{S}$ is a function of the current camera pose and the local map. To increase the number of map points in the central area of the FoV, we designed a scoring function (Eq. 7) based on $\cos({\alpha_{1}})$ (see Fig. \ref{FLAF} for $\alpha_{1}$) to rotate the camera toward regions dense with map points. By adding up the scores of all map points within the FoV, each one of them contributes to the total score of this PTU sample. Map points with smaller $\alpha_{1}$ values receive higher scores, which encourages the clustering of map points in the central area of the FoV. Consequently, $N_{S}$ is increased and tracking failure probability is decreased compared to passive VSLAM. This effect can be represented by the inequality:
\begin{equation}
    f(\mathbf{T}_{pt} \mathbf{X}_{k}) > f(\mathbf{X}_{k})
\end{equation}

Another goal of our design is to reward the map points with good feature identifiability relative to the current FoV. Feature identifiability refers to the ability of the feature algorithm to recognize the map point from a specific position and orientation. With a complete path map built, there exist more local map points to observe for localization. However, some of the map points can not be identified by the feature detector from an arbitrary position and orientation due to appearance changes\cite{b10}. This phenomenon can be summarized as “looking at points visible but not identifiable”, which results in a higher failure rate of repeating. 

To address this, we take the view angle ($\alpha_{2}$ in Fig. \ref{FLAF}) into consideration and assume the keyframes that observe the map point define a view angle range of successful identification. The mean viewing direction is represented as the feature normal, around which the view angle range is defined. This concept of feature normal was also used in \cite{b9,b12}. Mostegel et al.\cite{b26} justified using $\cos{\alpha_{2}}$ as the metric of the probability of feature identification to account for the observation of a feature from different viewpoints and view angles. Our scoring function, $\cos({\alpha_{1}})\cdot\cos({\alpha_{2}})$, multiplies two cosine functions to prioritize map points that score highly on both metrics.  

\subsection{Path Learning and Tracking with Active Camera}
During teaching, the path map and a robot trajectory are incrementally constructed by VSLAM. After teaching, a complete map consisting of plenty of 3D map points and a graph of keyframes are saved with corresponding PTU angles read from the PTU encoder. The learned trajectory for repeating is stored as a set of key robot poses $\big\{\mathbf{X}_{RK}\big\}$, each one of which is derived from the keyframe pose and corresponding PTU angles:
\begin{equation}
     \mathbf{X}_{R,k} =  \mathbf{T}_{pt,k}^{-1} \mathbf{X}_{k}
    \label{eq8}
\end{equation}
Where $\mathbf{X}_{R}$ denotes the robot pose and $\mathbf{X}_{R}$ is the camera pose.

Once repeating begins, the previously taught map is loaded, and the AC-SLAM, as shown in Fig. \ref{framework of active SLAM}, is performed to localize the robot with the mapping module closed. Meanwhile, the active camera autonomously adjusts its orientation according to real-time perception. The current robot pose during the repeating is computed from the current PTU angles and camera pose by Eq. \eqref{eq8}. Following this, we search for the closest robot pose in $\big\{\mathbf{X}_{R,i}\big\}$ ahead of the current robot pose $\mathbf{X}_{R,k}$ as the current reference robot pose $\mathbf{X}_{r}$. Finally, the pose error between $\mathbf{X}_{r}$ and $\mathbf{X}_{R,k}$ is processed by a PD controller\cite{b24}, denoted as $\boldsymbol{C}_{pd}$, to calculate the current velocity $\boldsymbol{\phi}_{k}$:
\begin{equation} 
\boldsymbol{\phi}_{k} = \boldsymbol{C}_{pd}(\mathbf{X}_{r}-\mathbf{X}_{R,k}) 
\label{PD} 
\end{equation}
To expedite the reference keyframe search, a window centered on the last reference keyframe is defined with a fixed width of 10 keyframes.

\begin{table*}
\caption{Comparison between Passive VT\&R and Active VT\&R with different View Planning Methods}
\vspace{-3mm}
\label{Table I}
\setlength{\tabcolsep}{5pt}

\begin{center}
\begin{tabular}{lc|cccc}
\hline
\multicolumn{1}{c}{\multirow{2}{*}{Paths}} & \multicolumn{1}{l|}{\multirow{2}{*}{Metrics}} & \multirow{2}{*}{Passive VT\&R} & \multirow{2}{*}{UDVP-based Active\cite{b17}} & \multirow{2}{*}{FLAF without Scoring\cite{b12}} & \multirow{2}{*}{\textbf{FLAF-based Active (Ours)}} \\
\multicolumn{1}{c}{}                       & \multicolumn{1}{l|}{}                         &                                &                                    &                                   &                                             \\ \hline
\multirow{3}{*}{Path1(15.08m)}             & CR(\%)                                        & 94.36(14.23m)                   & 62.08(9.362m)                       & 72.88(10.99m)                      & \textbf{100(15.08m)}                        \\
                                           & Time(s)                                       & -                              & 0.2036                             & 0.1742                            & 0.2976                                      \\
                                           & AP-RMSE(m)                                    & 0.4992                         & 0.3772                             & 0.2324                            & 0.5333                                      \\ \hline
\multirow{3}{*}{Path2(19.34m)}             & CR(\%)                                        & 100(19.43m)                    & 65.20(12.61m)                       & 87.8(16.99m)                       & \textbf{100(19.52m)}                        \\
                                           & Time(s)                                       & -                              & 0.1973                             & 0.1631                            & 0.3015                                      \\
                                           & AP-RMSE(m)                                    & 0.3573                         & 0.5171                             & 0.4326                            & 0.3058                                      \\ \hline
\multirow{3}{*}{Path3(29.906)}            & CR(\%)                                        & \usym{1F5F4}  & 78.98(23.62)                        & 93.06(27.83)                      & \textbf{100(39.08)}                        \\
                                           & Time(s)                                       & -                              & 0.2365                             & 0.1784                            & 0.3219                                      \\
                                           & AP-RMSE                                    & -                              & 0.7700                             & 0.9301                            & 1.185                                       \\ \hline
\multirow{3}{*}{Path4(19.391)}                   & CR(\%)                                        & \usym{1F5F4}           & 62.11(12.04)               & 52.77(10.232)              & \textbf{91.27(17.69)}                       \\
                                           & Time(s)                                       & -           & 0.3076               & 0.3394              & 0.5089                        \\ 
                                           & AP-RMSE                                    & -                              & 0.5085                                   & 0.4929              & 0.6265                        \\ \hline

\multicolumn{6}{l}{All the data in Table I are the average results of 10 repeated experiments. \enquote{\usym{1F5F4}} indicates failures in the teaching phase. The AP-RMSE data of Paths 3 }\\
\multicolumn{6}{l}{and 4 are relative and lack a definite scale because the ground truths are derived using VSLAM with images captured by a monocular camera.}\\
\multicolumn{6}{l}{\enquote{CR} means the average completion rate in the repeating stage of VT\&R. \enquote{Time} means the average time used by the sampling-based view planner. }
\vspace{-5mm}
\end{tabular}
\label{tab1}
\end{center}
\end{table*}

\begin{figure*}
  \centering
  \includegraphics[scale=0.3]{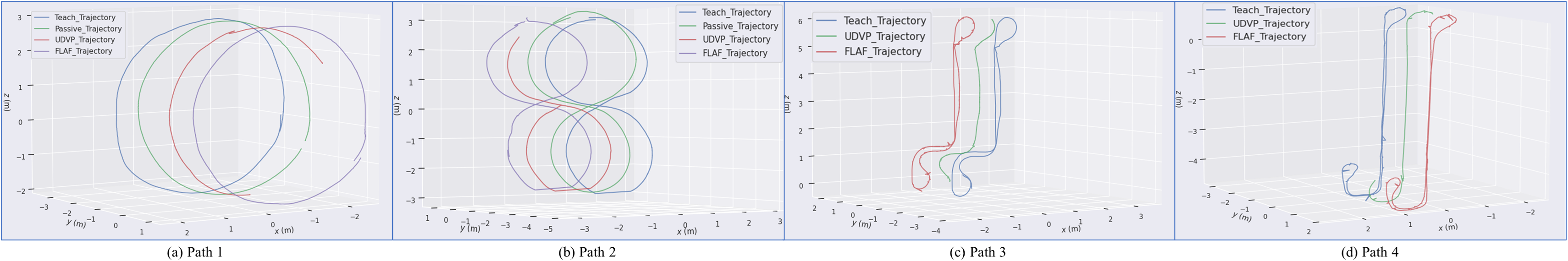}
  \vspace{-7mm}
  \caption{Trajectories of passive VT\&R and active VT\&R using different view planning methods are shown. Offsets on the Y-Axis are manually added to separate the overlapping teaching and repeating trajectories for better visualization. The teaching trajectories for Path 1 and Path 2 were obtained via \enquote{motion capture} as ground truth, while those for Paths 3 and Path 4 were derived using VSLAM. Our FLAF-based active VT\&R system demonstrates the highest completion rate (CR) across all four paths and can reliably navigate all four paths over multiple loops with very few failures.}
  \label{trajectories}
  \vspace{-6mm}
\end{figure*}

\section{Experiments and Discussion}

Our experiments are primarily designed to demonstrate our repeatable and successful VT\&R on challenging paths. As the trajectory errors of repeating are all acceptable, we emphasize the completion rate (CR) and success rate (SR), which indicates that only our FLAF-based active VT\&R can complete all four paths at a high SR.

\subsection{Implementation and Experimental Setup}\label{AA}

As shown in Figure \ref{fig1}, we fix an Intel Realsense D435 camera on an I-Quotient-Robotics PTU to make up our active camera, which is mounted on a Clearpath-Jackal robot. Our active VT\&R system operates in real-time on a notebook computer equipped with an Intel i7(2.3GHz) processor and responds to the images exactly at the frame rate of 20Hz.

Experiments are performed on 4 paths to evaluate our view planner and VT\&R system. The first two paths are in the effective range of the motion capture device in the Shenzhen Key Laboratory of Robotics and Computer Vision. Paths 3 and 4 respectively lead the robot from inside the laboratory to a space out of the laboratory and finally back to the start. All the data in Table I are the average results of 10 repeated experiments. The trajectories shown in Fig. \ref{trajectories}, the map shown in Fig. \ref{realscenario}, and the plots shown in Fig. \ref{curve} are a representative selection, considering that the previous methods consistently fail at a similar location across multiple repetitions.

On Paths 1 and 2, we demonstrate the efficacy of our VT\&R system in both an active and a passive way. On Paths 3 (Fig. \ref{realscenario}) and 4, we show challenging cases with low-texture regions where passive VT\&R fails and active VT\&R succeeds. Additionally, our FLAF view planner is verified on all four paths to outperform the existing UDVP in repeating a complete path. FLAF without scoring refers to counting the map points in a range defined by FLAF instead of grading the points by the product of the cosine functions shown in Equation \eqref{active optimization}.

\subsection{Tracking Failure Avoidance Validation}\label{SCM}

As in Table \ref{tab1} and Fig. \ref{trajectories}, the passive VT\&R achieves a stable and accurate performance on the first two paths but fails in the teaching phase on Paths 3 and 4. The few low-texture regions on Paths 1 and 2 are avoided by a considerable human guide. Our active VT\&R system with three view planners succeeds in the teaching phase on all 4 paths.

\begin{figure}
\vspace{-1mm}
  \centering
  \includegraphics[scale=0.195]{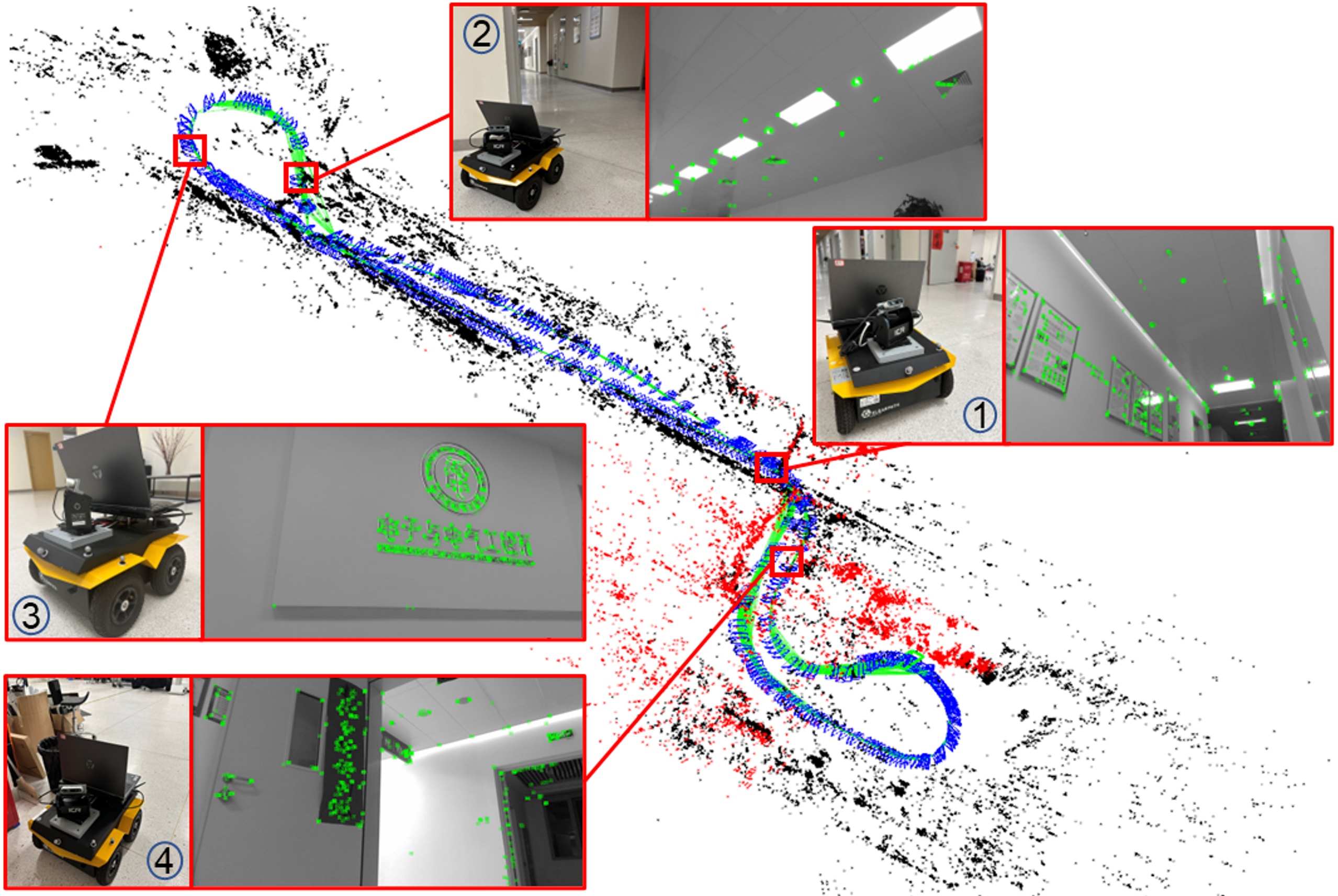}
  \vspace{-7mm}
  \caption{Feature map of Path 3 built by our active SLAM and robot views in repeating phase at challenging locations.}
  \label{realscenario}
  \vspace{-5mm}
\end{figure}

On Path 3, which connects several rooms, Fig. \ref{realscenario} illustrates how our active VT\&R successfully navigates challenging low-texture regions. The active camera autonomously focuses on informative areas, ensuring stable localization throughout. At position 1, the active camera orients toward the poster in the upper left to avoid the white wall. At position 2, the active camera looks up at the ceiling to maintain the localization relying on the square lamps. At position 3, the robot orients toward the upper right to focus on the logo while passing through a low-texture corner. Finally, at position 4, the robot looks toward the upper left at the door for abundant features. 

\begin{table}[]

\scriptsize
\caption{Success Rate (SR) analysis of the VT\&R system in different view planning methods.}
\vspace{-1mm}
\setlength{\tabcolsep}{2pt}
\begin{tabular}{ccccc}
\hline
Methods                    & Path1(15.1m) & Path2(19.3m) & Path3(29.9) & Path4(19.4) \\ \hline
Passive VT\&R              & 100\%              & 100\%         & \usym{1F5F4}            & \usym{1F5F4}             \\
UDVP-based Active          & 46.7\%             & 13.3\%        & 20\%              &  0\%             \\
FLAF without Scoring           & 73.3\%             & 66.7\%        & 53.3\%              & 33.3\%              \\
\textbf{FLAF-based Active} & \textbf{100\%}        & \textbf{100\%}         & \textbf{100\%}         & \textbf{73.3\%}        \\ \hline
\multicolumn{5}{l}{ \enquote{SR} indicates the success rate of completing the entire path in repeating.} \\
\multicolumn{4}{l}{ The SR data in Table II are the results of 15 repeated experiments. } \\
\multicolumn{5}{l}{ \enquote{\usym{1F5F4}} indicates failure in teaching. } \\
\end{tabular}

\vspace{-7.5mm}
\end{table}

\subsection{Active View Planning Method Comparison}
We compared our FLAF-based view planner against three other methods: Passive VT\&R, UDVP-based active VT\&R, and FLAF without scoring-based active VT\&R. The passive VT\&R is achieved on our VT\&R system without incorporating active view planning. We implement the UDVP-based active VT\&R  by reproducing the view planner proposed in \cite{b17} with our VT\&R framework. The comparison with \enquote{FLAF without scoring} serves as an ablation study, illustrating the impact of our scoring mechanism as described by equation \eqref{active optimization}. FLAF without scoring-based active VT\&R can also be seen as an active VT\&R with the model depicted in \cite{b12}. Ground truth of trajectories for Paths 1 and 2 were collected using motion capture during the teaching phase, while for Path 3, VSLAM was used to generate the ground truth, as motion capture was unavailable outside the laboratory.

In Table I, we compare the view planning methods in three metrics: (1) \enquote{CR} indicates the VT\&R completion rate using different view planning methods, (2) \enquote{Time} indicates the average time used for the view planning methods implemented by sampling-based optimization, and (3) \enquote{AP-RMSE} indicates the repeating precision by absolute pose-RMSE, which is computed using evo\cite{b25} by comparing the repeating trajectory with the taught one based on timestamps, resulting in error data greater than the actual situation. Our FLAF-constrained view planner outperforms the UDVP method in repeating complete paths because it accounts for the affine change of feature points. The normal line of the map point ($\mathbf{n}_{p}$), shown in Fig. \ref{FLAF}, limits the orientation of the active camera in the view angle-invariant range of the feature. The efficacy of VSLAM is decreased by the relatively low speed of view planning compared to the SLAM speed of 20Hz.

In Fig. \ref{trajectories}, we visualize the trajectories of active VT\&R with different view planners, which confirms that the repeated trajectories with different methods all align well with the taught ones. Although the UDVP method achieves more accurate path following on Path 3, the trajectory errors on all paths by all view planners are negligible for VT\&R. A portion of the AP-RMSE arises from point-to-point comparisons over time, which are difficult to execute precisely. To address this, we adjusted the timestamps to ensure uniform consistency over the same time duration.

\subsection{Map Points Association Validation}

In\cite{b12}, the authors present a line graph illustrating the relationship between the probability of the tracking failure and the number of observed feature points. Their results\cite{b12} indicate that the likelihood of the tracking failure approaches zero when the number of associated map points exceeds a certain threshold. From the perspective of our work, the specific failure is caused by choosing the wrong orientation of the active camera relative to the local map.  To further investigate, we applied the analysis method proposed by \cite{b12} to examine the state of map point association during the repeating phase. 

\begin{figure}
  \centering
  \includegraphics[scale=0.182]{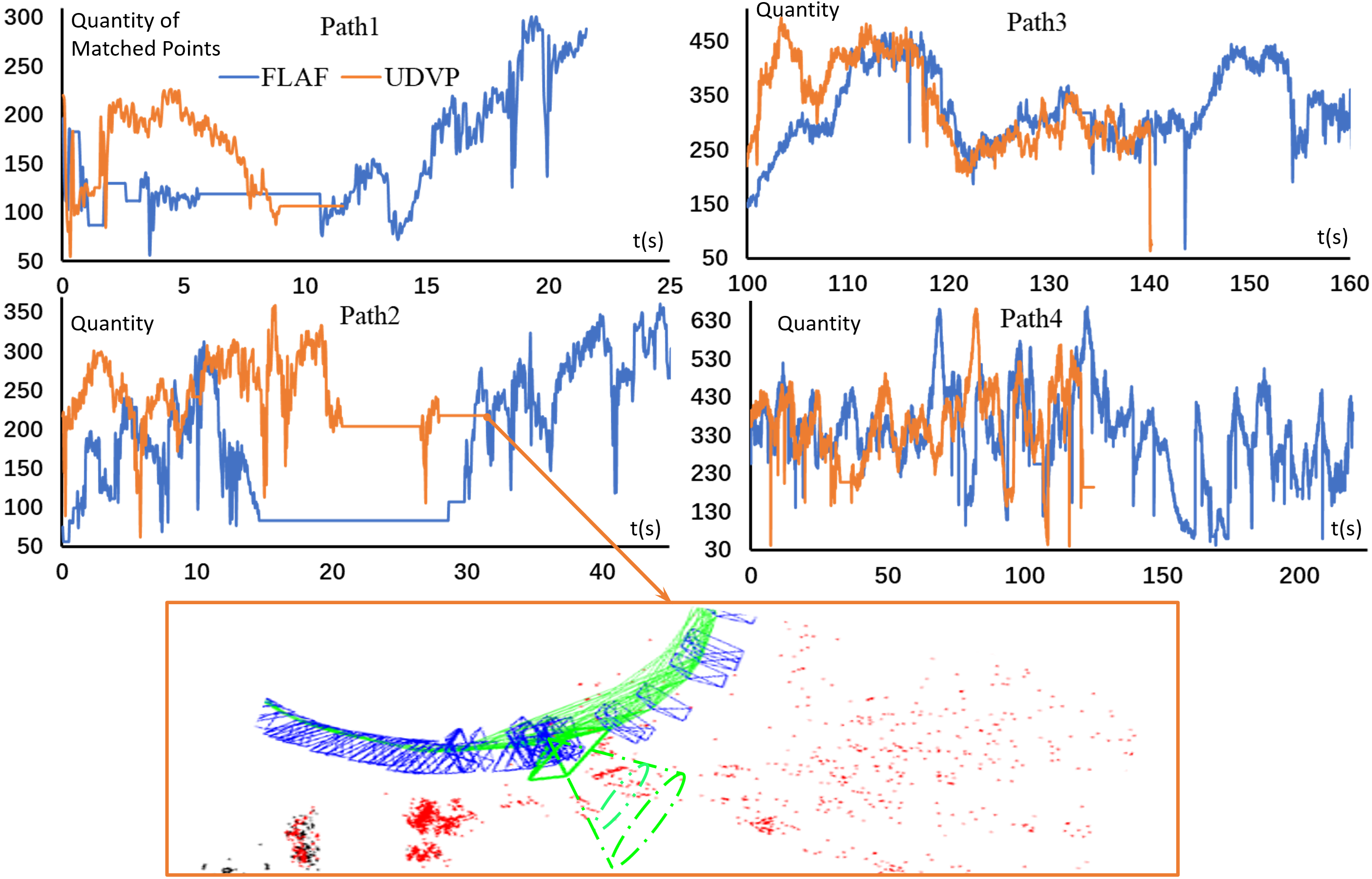}
  \caption{Number of matched points in local map tracking with respect to time. The bottom shows the failure case of the UDVP method on Path 2 shown in a feature map. The green square represents the camera's current pose.}
  \label{curve}
  \vspace{-6mm}
\end{figure}

As shown in Fig. \ref{curve}, we recorded the number of inlier points successfully matched during local map tracking and plotted line graphs comparing the performance of the two methods across all test paths. On Paths 1 and 2, the active camera-based VSLAM with FLAF initially matched fewer points than the UDVP-based system. However, after an initial reduction of matched points, the UDVP-based method fails to maintain tracking, while the FLAF-based method continues to provide stable localization and an increasing number of matches. This phenomenon suggests that, once the inlier number of the tracked map points reaches the threshold, the sheer number of map points becomes less critical for stable localization. The bottom of Fig. \ref{curve} also illustrates how the UDVP method directs the active camera toward regions with more local map points, without accounting for the feature identifiability. Although many points may fall within the camera's FoV, they may not be recognized or matched by the feature extractor due to the ignorance of the affine changes. 

On Paths 3 and 4, the active visual repeat with both the FLAF and UDVP view planners tracks a similar quantity of map points. However, our FLAF-based method successfully recovers localization after a challenging decline in tracked points, where the UDVP-based approach fails. Even in cases of temporary tracking loss, the active camera controlled by our view planner was able to orient itself appropriately by executing the instruction before tracking loss, allowing the VSLAM system to recover localization through place recognition and the PnP algorithm.

\section{Conclusion}

In this research, we present a novel active view planning method for VT\&R that addresses the tracking failure caused by the low-texture regions and demonstrates the whole active VT\&R. Our experimental results show that our active VT\&R successfully overcame the specific failure of passive VT\&R and our proposed FALF-constrained active view planning outperforms existing view planners in completion and success rate of VT\&R. 

During our tests, the VT\&R systems built on existing view planners frequently failed in the repeat phase without considering the feature-identifiability of the map points. Our proposed focal line and feature (FLAF)-constrained active view planning successfully addressed these failures by considering the view angle difference between the current viewpoint and those at which the map points were triangulated. With our view planner, the active VT\&R system successfully finishes all four paths at the highest completion and success rate. Additionally, our point-line plots indicate that the quality of the map points is more important than quantity for stable localization.

However, for each execution of view planning, 900 samples of PTU angles were scored according to thousands of map points in the local map, resulting in a high computational overhead. Despite using OpenMP to speed up the view planners, they operated at less than 5 FPS, which hindered the performance of VSLAM and reduced the success rate of VT\&R by preventing accurate path reconstruction. In future work, we will address this limitation by parallelizing the view planning module with the VSLAM system to improve processing speed and overall system performance.

\vspace{12pt}


\begin{thebibliography}{00}
\vspace{-2mm}
\bibitem{b1} Paul Furgale, and Timothy D Barfoot, \enquote{Visual teach and repeat for long-range rover autonomy,} Journal of Field Robotics, 2010.

\bibitem{b2} Michael Paton, Kirk MacTavish, Michael Warren, Timothy D Barfoot, \enquote{Bridging the appearance gap: Multi-experience localization for long-term visual teach and repeat.} In IEEE/RSJ Int. Conf. on Intell. Robots and Systems, 2016.

\bibitem{b3} L. Peterson, D. Austin, and D. Kragic, \enquote{High-level control of a mobile manipulator for door opening,} In IEEE/RSJ Int. Conf. on Intell. Robots and Systems, 2000.

\bibitem{b4} Mohit Mehndiratta and Erdal Kayacan, \enquote{A constrained instantaneous learning approach for aerial package delivery robots: onboard implementation and experimental results,} Autonomous Robots, 2019.

\bibitem{b5} Guillaume Bresson, Zayed Alsayed, Li Yu, and Sébastien Glaser, \enquote{Simultaneous Localization and Mapping: A Survey of Current Trends in Autonomous Driving,} IEEE Trans. Intell. Vehicles, 2017.

\bibitem{b6} H Ye, G Chen, W Chen, L He, Y Guan, and H Zhang, \enquote{Mapping While Following: 2D LiDAR SLAM in Indoor Dynamic Environments with a Person Tracker,} In IEEE Int. Conf. on Robot. and Biomimetics, 2021.


\bibitem{b7} Jakob Engel, Vladlen Koltun, and Daniel Cremers. \enquote{Direct sparse odometry.} IEEE Trans. Pattern Anal. Mach. Intell, 2017.

\bibitem{b8} Christian Forster, Matia Pizzoli, and Davide Scaramuzza. \enquote{SVO: Fast semi-direct monocular visual odometry.} In IEEE Int. Conf. on Robotics and Automation, 2014.


\bibitem{b9} R. Mur-Artal, J. M. M. Montiel, and J. D. Tardos, \enquote{Orb-slam: a versatile and accurate monocular slam system,} IEEE Trans. Robotics, 2015.

\bibitem{b10}David G Lowe. \enquote{Distinctive image features from scale-invariant keypoints.} International journal of Computer Vision, 2004.

\bibitem{b11} Seyed Abbas Sadat, Kyle Chutskoff, Damir Jungic, Jens Wawerla and Richard Vaughan, \enquote{Feature-Rich Path Planning for Robust Navigation of MAVs with Mono-SLAM,} In IEEE Int. Conf. on Robotics and Automation, 2014.

\bibitem{b12} Xinke Deng, Zixu Zhang, Avishai Sintov, Jing Huang, and Timothy Bretl, \enquote{Feature-constrained Active VSLAM for Mobile Robot Navigation,} In IEEE Int. Conf. on Robotics and Automation, 2018.



\bibitem{b14} Matias Mattamala, Milad Ramezani, Marco Camurri, and Maurice Fallon. \enquote{Learning Camera Performance Models for Active Multi-camera Visual Teach and Repeat.} In IEEE Int. Conf. Robotics and Automation, 2021.

\bibitem{b15} Simone Frintrop, and Patric Jensfelt. \enquote{Attentional Landmarks and Active Gaze Control for VSLAM.} IEEE Trans. Robotics, 2008.

\bibitem{b16} Andrew J. Davison and David W. Murray. \enquote{Mobile Robot Localisation using Active Vision.} In European Conf. Computer Vision, 1998.

\bibitem{b17} Xu-Yang Dai, Qing-Hao Meng, and Sheng Jin. \enquote{Uncertainty-driven Active View Planning in Feature-based Monocular vSLAM.} Applied Soft Computing, 2021.


\bibitem{b18} Michael Warren, Angela P. Schoellig, and Timothy D. Barfoot. \enquote{Level-headed: Evaluating Gimbal-stabilised Visual Teach and Repeat for Improved Localisation performance.} In IEEE Int. Conf. Robotics and Automation, 2018.

\bibitem{b19} Hauke Strasdat, José MM Montiel, and Andrew J. Davison. \enquote{VSLAM: why filter?} Image and Vision Computing, 2012.


\bibitem{b20} Weinan Chen, Changfei Fu, Lei Zhu, Shing-Yan Loo, and Hong Zhang. \enquote{Rumination Meets VSLAM: You Do Not Need to Build All the Submaps in Realtime.} IEEE Trans. Industrial Electronics, 2023.

\bibitem{b21} Etienne Mouragnon, Maxime Lhuillier, Michel Dhome, Fabien Dekeyser, and Patrick Sayd.\enquote{Real Time Localization and 3D Reconstruction.} In IEEE Int. Conf. Computer Vision and Pattern Recognition, 2006.


\bibitem{b22} Ethan Rublee, Vincent Rabaud, Kurt Konolige, and Gary Bradski. \enquote{ORB: An efficient alternative to SIFT or SURF.} In IEEE Int. Conf. Computer Vision, 2011.

\bibitem{b23} Vincent Lepetit, Francesc Moreno-Noguer, and Pascal Fua. \enquote{EPnP: An accurate O(n) Solution to the PnP Problem.} International Journal of Computer Vision, 2009.

\bibitem{b24} Weinan Chen, Lei Zhu, Xubin Lin, Li He, Yisheng Guan, and Hong Zhang. \enquote{Dynamic Strategy of Keyframe Selection with PD Controller for VSLAM Systems.} IEEE/ASME Trans. Mechatronics, 2021.

\bibitem{b25} Michael Grupp. \enquote{evo: Python package for the evaluation of odometry and slam.} 2017, Available: https://github.com/MichaelGrupp/evo.

\bibitem{b26} Christian Mostegel, Andreas Wendel, and Horst Bischof. \enquote{Active Monocular Localization: Towards Autonomous Monocular Exploration for Multirotor MAVs.} In IEEE Int. Conf. Robotics and Automation, 2014.

\end{thebibliography}
\end{document}